\UseRawInputEncoding
\documentclass[letterpaper, 10 pt, conference]{ieeeconf}  

\IEEEoverridecommandlockouts                              

\usepackage{amsmath} 
\usepackage{amssymb}  
\usepackage{soul}
\usepackage{tikz}
\usepackage{mpac_tikz_tools}
\usepackage{array}
\usepackage{caption}
\usepackage{subcaption}
\usepackage{algorithm, algpseudocode}  
\newcommand{\WRP}{~\dots\par\hspace{30mm}\qquad\enspace}
\let\proof\relax  
\usepackage{amsthm}
\theoremstyle{definition}

\newtheorem{theorem}{\normalfont\bfseries Theorem}

\newtheorem{definition}{\normalfont\bfseries Definition}

\newtheorem{example}{\normalfont\bfseries Example}

\newtheorem{manualtheoreminner}{}
\newenvironment{primitive}[1]{%
  \renewcommand\themanualtheoreminner{#1}%
  \manualtheoreminner
}{\endmanualtheoreminner}

\newcommand{\motionprimitive}{\mathcal{P}}
\newcommand{\transitions}{\mathcal{T}}

\newcommand{\textbfit}[1]{\textbf{\textit{{#1}}}}
\newcommand{\motGraph}{BuildMotionPrimitiveGraph}
\usepackage{environ}

\NewEnviron{ruledtable}{%
\par\addvspace{2mm}\hrule height 0.03cm 
\begin{table}[!h]\BODY\end{table}
\hrule height 0.03cm \addvspace{2mm}
}

\title{\LARGE \bf
Verifying Safe Transitions between Dynamic Motion Primitives on Legged Robots
}

\author{Wyatt Ubellacker, Noel Csomay-Shanklin, Tamas G. Molnar, and Aaron D. Ames%
        \thanks{This research is supported by DOW Chemical, project 227027AT.}%
\thanks{Authors are with the Department of Control and Dynamical Systems and the Department of Mechanical and Civil Engineering, California Institute of Technology, Pasadena, CA, USA.
{\tt\small wubellac, noelcs,tmolnar,ames@caltech.edu }}
}

\begin{document}

\maketitle
\thispagestyle{empty}
\pagestyle{empty}

\begin{abstract}

Functional autonomous systems often realize complex tasks by utilizing state machines comprised of discrete primitive behaviors and transitions between these behaviors.  This architecture has been widely studied in the context of quasi-static and dynamics-independent systems.  However, applications of this concept to dynamical systems are relatively sparse, despite extensive research on individual dynamic primitive behaviors, which we refer to as ``motion primitives.'' This paper formalizes a process to determine dynamic-state aware conditions for transitions between motion primitives in the context of safety.  The result is framed as a ``motion primitive graph'' that can be traversed by standard graph search and planning algorithms to realize functional autonomy.  To demonstrate this framework, dynamic motion primitives---including standing up, walking, and jumping---and the transitions between these behaviors are experimentally realized on a quadrupedal robot.  
\end{abstract}

\section{INTRODUCTION}

There is a significant volume of literature, applications, and demonstrations of functional
autonomous systems that perform a wide range of complex tasks, ranging from the unstructured
environment of disaster-response motivated DARPA Robotics Challenge \cite{DARPA} to
the highly structured environment of industrial manufacturing robotics \cite{PEDERSEN2016282}. These
systems often use sequence-based \cite{TeamRobosimianKaruma2017,edelbergCASAH}, state-machine
based \cite{state_mach}, or graph-based \cite{kim2001executing} autonomy to couple many
discrete behaviors together to complete highly complex tasks.
These applications are typically quasi-static or have dynamics which are self-contained within each primitive behavior; thus, task-level or pose states can be used to determine if transitions are safe, while dynamic states are not considered. 
If one intends to extend this concept to dynamic systems and transitions,
then the dynamic nature of the primitive behaviors and the transitions between them must be considered explicitly. This observation motivates the desire to determine safe
transitions between what are referred to as ``motion primitives''
\cite{HumanInspiredZhao2014,motion_prim_Paranjape,motion_prim_Kolathaya}.

There is a wealth of prior work on motion primitives, examples of which include walking,
running and jumping for legged robotics \cite{westervelt2018feedback,kim2019highly},
lane-following, cruise control and parallel parking for wheeled vehicles
\cite{gomez2001parallel}, and hovering and landing for quadrotor applications
\cite{falanga2017vision}.  It is important to note that there exist many different methods to
realize a given behavior, with varying advantages and disadvantages in the aspects of safety,
performance, robustness to uncertainty, etc.
Certain methods for generating motion primitives may be more or less
conducive for use in our proposed framework.
Rather than focus on specific approaches, the goal of this paper is to detail a general framework for studying motion primitives and transitions that is verifiable and, importantly, realizable on dynamic legged robotic systems. 

Of special interest in this work are transitions between dynamic motion primitives.
For these primitives, transitioning requires convergence to the next primitive while maintaining safety, and can be studied through the notions of stability and regions of attraction.
There are a number of effective techniques to estimate regions of attraction, including backward reachability methods \cite{Yuan2019}, optimization-based Lyapunov methods \cite{MATALLANA2010574, johansen2000computation, polanski1997lyapunov}, quadratic Lyapunov
methods \cite{davison1971computational} and linearization methods \cite{khalil2002nonlinear}.
Additionally, Control Lyapunov Function and Control Barrier Function-based motion
primitives \cite{ames2016control} can also be used to enforce both stability and safety explicitly.
In this paper, we leverage these ideas to verify safe transitions.

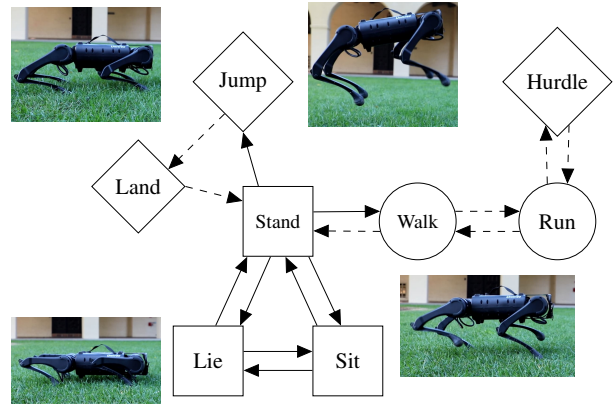
\begin{figure}
    \centering
            \resizebox{0.925\columnwidth}{!}{%
    \begin{tikzpicture}
      \node[fixed node] (Stand) at (0,0) {\footnotesize Stand};
      \node[periodic node] (Walk) at (2,0) {\footnotesize Walk};
      \transition[class1](Stand,Walk)();
      \transition[class2](Walk,Stand)();

      \node[fixed node] (Sit) at (1,-2) {\small Sit};
      \node[fixed node] (Lie) at (-1,-2) {\small Lie};
      \node[periodic node] (Run) at (4,0) {\small Run};
      \transition[class1](Stand,Lie)();
      \transition[class1](Stand,Sit)();
      \transition[class1](Sit,Stand)();
      \transition[class1](Sit,Lie)();
      \transition[class1](Lie,Stand)();
      \transition[class1](Lie,Sit)();
      \transition[class2](Walk,Run)();
      \transition[class2](Run,Walk)();

      \node[transient node] (Land) at (-2,0.5) {\small Land};
      \node[transient node] (Jump) at (-0.5,2) {\small Jump};
      \node[transient node] (Hurdle) at (4,2) {\small Hurdle};
      \transition[class1](Stand,Jump)();
      \transition[class2](Jump,Land)();
      \transition[class2](Land,Stand)();
      \transition[class2](Run,Hurdle)();
      \transition[class2](Hurdle,Run)();
    \node[inner sep=0pt] (LiePic) at (-2.75,-2)
    {\includegraphics[width=.12\textwidth]{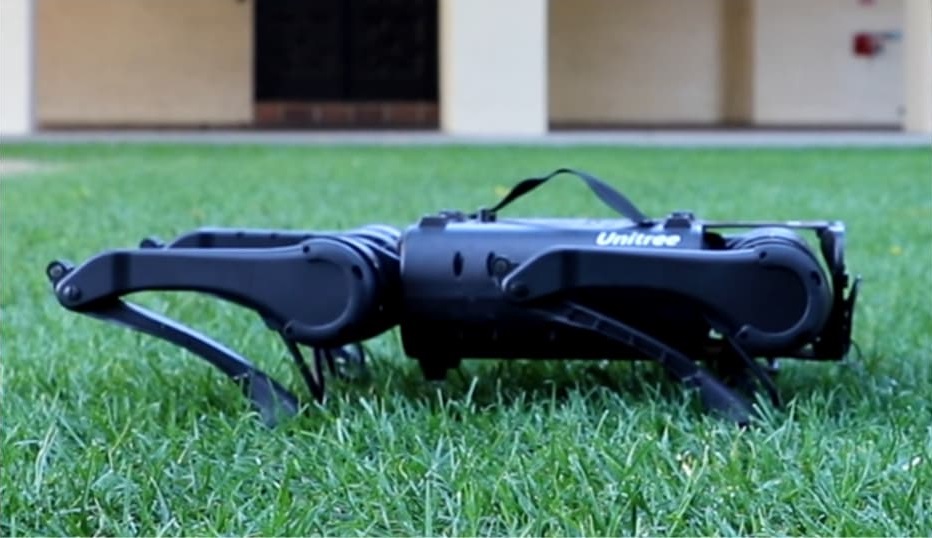}};
    \node[inner sep=0pt] (WalkPic) at (3,-1.5)
    {\includegraphics[width=.14\textwidth]{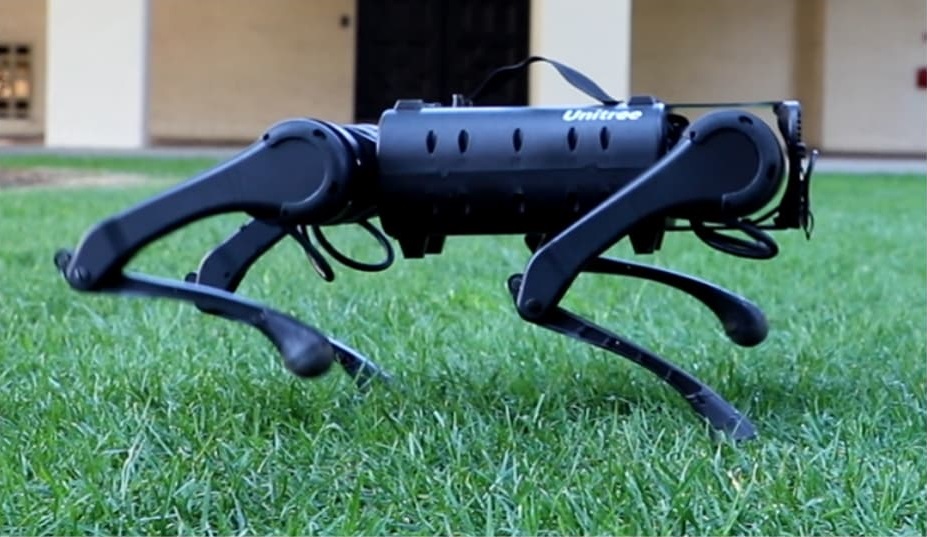}};
    \node[inner sep=0pt] (JumpPic) at (1.5,2.25)
    {\includegraphics[width=.12\textwidth]{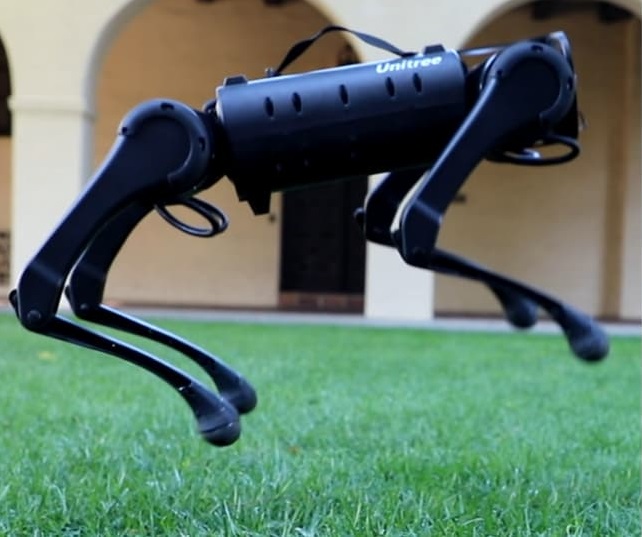}};
    \node[inner sep=0pt] (LandPic) at (-2.75,2.25)
    {\includegraphics[width=.12\textwidth]{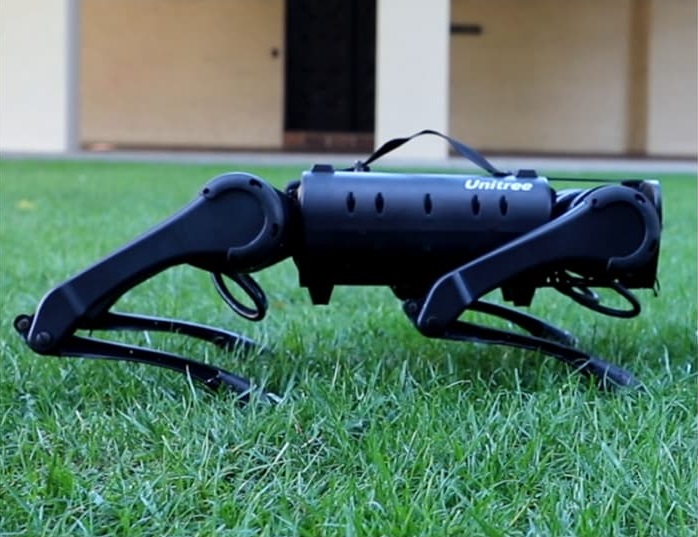}};
    \end{tikzpicture}} \\[4pt]
        \caption{A depiction of the motion primitive graph and the Unitree A1 executing the graph dynamically.}
    \label{quad_graph}
    \vspace{-5mm}
\end{figure}

The main result of this paper is a formalization of dynamic motion primitives and transition
types between them, and a method to verify the availability of transitions between arbitrary
primitives. This leads to the construction of a ``motion primitive graph'' that can be utilized
via standard graph search algorithms to provide a certifiably-safe transition path between the
execution of motion primitives; see Figure~\ref{quad_graph}. This method was applied to a set of quadrupedal motion primitives and validated through hardware experiments on a Unitree A1 quadruped. The results demonstrate successful and safe execution of a complex sequence of desired dynamic behaviors, e.g., walking and jumping, and a comparison with na\"ive transitioning highlights the contributions of this work.



\section{Preliminaries}
\label{sec:preliminaries}

Consider a nonlinear system in control affine form:
\begin{align*}
    \dot x = f(x) + g(x) u
\end{align*}
where ${x \in \mathcal{X}\subset\mathbb{R}^n}$ is the state of the system, $u \in \mathcal{U}\subset\mathbb{R}^m$ represents the control inputs, and the functions ${f:\mathcal{X}\to \mathbb{R}^n}$ and ${g:\mathcal{X}\to \mathbb{R}^{n\times m}}$ are assumed to be locally Lipschitz continuous. Under the application of
a locally Lipschitz continuous feedback control law ${u=k(x,t)}$, we get the closed loop dynamics:
\begin{align}
    \dot{x}
    = f_{\rm cl}(x,t)
    = f(x) + g(x) k(x,t). \label{eq:CL_dynamics}
\end{align}
For these dynamics, the solution to the initial value problem
with ${x(0)=x_0}$ is termed the \textit{flow} of the system and is denoted as $\phi_t(x_0)$.
The flow exists for all $t \geq 0$ if we further assume that $\mathcal{X}$ and $\mathcal{U}$ are compact.

In the following constructions, we describe the closed-loop behavior $\phi_t(x_0)$ by the help of \textit{motion primitives}.
While motion primitives are not a novel concept \cite{HumanInspiredZhao2014,motion_prim_Paranjape,motion_prim_Kolathaya}, we
introduce our own definition that facilitates
the construction of motion primitive graphs for use in planning and autonomy.

\begin{definition}{}
\label{def:primitive}
A \textbfit{motion primitive} is a dynamic behavior of~(\ref{eq:CL_dynamics})
defined by the 6-tuple $\motionprimitive={(x^*,k,\Omega,\mathcal{C},\mathcal{S},\mathcal{E})}$ with the following attributes.
\begin{itemize}
\item The \textit{setpoint}, ${x^*:\mathbb{R} \to \mathcal{X}}$, that describes the desired state as a function of time by $x^*(t)$.
It satisfies~(\ref{eq:CL_dynamics}) and hence ${x^*(t+t_0)=\phi_t(x^*(t_0))}$, ${\forall t \geq 0, t_0 \in \mathbb{R}}$.
It may be executed with a selected initial time $t_0$.
\item The \textit{control law}, ${k:\mathcal{X} \times \mathbb{R} \to \mathcal{U}}$, that determines the control input ${u=k(x,t)}$.
It is assumed to render the setpoint locally asymptotically stable.
\item The \textit{region of attraction (RoA)} of the setpoint, ${\Omega: \mathbb{R} \to \mathcal{X}}$, given by ${\Omega(t_0) \subseteq \mathcal{X}}$:
\begin{align}
    \Omega(t_0) = \{x_0\in\mathcal{X} : \lim_{t\to \infty} \phi_t(x_0) \!-\! x^*(t\!+\!t_0) = 0\}.
    \label{eq:RoA}
\end{align}

\item The \textit{safe set}, ${\mathcal{C}: \mathbb{R} \to \mathcal{X}}$, that indicates the states of safe operation by the set ${\mathcal{C}(t) \subset \mathcal{X}}$.
It is assumed to be the 0-superlevel set of a function
${h:\mathcal{X} \times \mathbb{R} \to \mathbb{R}}$:
\begin{equation*}
\mathcal{C}(t)=\{ x \in \mathcal{X} : h(x,t) \geq 0 \},
\end{equation*}
such that ${x^*(t+t_0) \in \mathcal{C}(t)}$, ${\forall t \geq 0, t_0 \in \mathbb{R}}$.
\item The \textit{implicit safe region of attraction}, ${\mathcal{S}: \mathbb{R} \to \mathcal{X}}$, that defines
the set of points from which the flow converges to the setpoint while being safe for all time:
\begin{equation}
\mathcal{S}(t_0)=\{x_0 \in \Omega(t_0): \phi_t(x_0) \in \mathcal{C}(t), \forall t \geq 0\}.
\label{eq:safeRoA}
\end{equation}
This set is non-empty as ${x^*(t_0) \in \mathcal{S}(t_0)}$.
In general, it is difficult to compute this set, thereby the term implicit.
\item The \textit{explicit safe region of attraction}, ${\mathcal{E}: \mathbb{R} \to \mathcal{X}}$.
Because finding $\mathcal{S}(t_0)$ is difficult, we consider a smaller, known subset ${\mathcal{E}(t_0) \subseteq \mathcal{S}(t_0)}$ with ${x^*(t_0) \in \mathcal{E}(t_0)}$, ${\forall t_0 \in \mathbb{R}}$.
\end{itemize}
\end{definition}



\subsection{Motion Primitive Classes}

We consider three classes of motion primitives, \textit{fixed}, \textit{periodic} and \textit{transient}, given by the examples below.

\begin{example}{}
A \textit{fixed motion primitive} is a behavior
where the setpoint and safe RoA are constant in time: ${x^*(t) = x^*}$, ${\mathcal{S}(t) = \mathcal{S}}$, ${\forall t \geq 0}$.
Examples in this class include stand and sit behaviors for a legged robots, hover in place for a multirotor, and hold position behavior for a robotic manipulator.
\end{example}

\begin{example}{}
A \textit{periodic motion primitive} is a behavior
where the setpoint and safe RoA are periodic in time: ${x^*(t) =
x^*(t+T_{\rm p})}$, ${\mathcal{S}(t) = \mathcal{S}(t+T_{\rm p})}$, with period ${T_{\rm p}>0}$.
Examples of this class are walking or running for legged robotics and repeated motions for pick-and-place
robotics.
\end{example}

\begin{example}{}
A \textit{transient motion primitive} is a behavior
where the setpoint and safe RoA are functions of time
on a finite domain: $x^*(t)$, $\mathcal{S}(t)$, ${t \in [t_0,t_{\rm f}]}$.
Jumping and landing behaviors on legged platforms and tracking path-planned trajectories are examples of transient motion
primitives.
\end{example}

These motion primitive classes are summarized in Table~\ref{table_mp_classes}.
For each class indicators will be used for visualization in motion primitive graphs:
box denotes fixed, circle represents periodic and diamond indicates transient motion primitives.

\begin{table}
\caption{Motion Primitive Classes}
\vspace{-3mm}
\label{table_mp_classes}
\begin{center}
\begin{tabular}{|c|c|c|c|}
\hline
Name & Setpoint & Safe RoA & Indicator \\
\hline
Fixed & $x^*(t) = x^*$  & $\mathcal{S}(t) = \mathcal{S}$ & 
\begin{tikzpicture}[baseline=0]\node[fixed node,scale=0.25] at (0,0.1){};\end{tikzpicture}\\
\hline
Periodic & $x^*(t) = x^*(t+T)$  & $\mathcal{S}(t) = \mathcal{S}(t+T)$ &
\begin{tikzpicture}[baseline=0]\node[periodic node,scale=0.25] at (0,0.1){};\end{tikzpicture}\\
\hline
Transient & $x^*(t), t\in[t_0,t_{\rm f}]$  & $\mathcal{S}(t), t\in[t_0,t_{\rm f}]$ &
\begin{tikzpicture}[baseline=0]\node[transient node,scale=0.25] at (0,0.1){};\end{tikzpicture}\\
\hline
\end{tabular}
\end{center}
\vspace{-6.5mm}
\end{table}

\subsection{Motion Primitive Transitions}

Having defined motion primitives,
we investigate the transitions
between them.
Let us consider motion primitives $A,B\in \motionprimitive$, a time moment $t_A\in \mathbb{R}_{\geq 0}$, the setpoint $x_A^*$ of $A$ and the controller $k_B$ of $B$.

\begin{definition}{}
\label{def:transition}
The flow $\phi_t^B(x_A^*(t_A))$ under controller $k_B$ starting from $x_A^*(t_A)$ is called a \textbfit{transition} $\transitions_{AB}$ from $A$ to $B$ starting at time $t_A$,
if ${\exists t_B \in \mathbb{R} \;\; \text{s.t.}\;\; x_A^*(t_A) \in \mathcal{S}_B(t_B)}$.
\end{definition}

To achieve a transition from $A$ to $B$, the control law $k_B$ of primitive $B$ can be applied starting from $t_A$ by selecting the appropriate $t_B$.
The transition is safe by construction, since the definition of $S_B$ implies ${\phi_t^B(x_A^*(t_A)) \in \mathcal{C}(t)}$, ${\forall t \geq t_A}$.
If $t_B$ exists for all
${t_A \geq 0}$,
the transition can be initiated at anytime from primitive $A$.
We call this case as \textit{Class 1 transition}.
Otherwise, transition can only be initiated at some specific times $t_A$, which is called \textit{Class 2 transition}.
Class 1 and Class 2 transitions will be represented by solid and dashed edges in the motion primitive graph, respectively.
If a transition does not exist between two primitives, then the control law $k_B$ may not guarantee a safe transition and no edge will exist in the motion primitive graph.
%
In what follows, we reduce the problem of transitioning safely to verifying the existence of Class 1 or 2 transitions and switching from the controller of primitive $A$ to that of $B$ at the right moment of time.

\subsection{Motion Primitive Graph}

Information about motion primitives and transitions between them can be organized into motion primitive graphs.

\begin{definition}
A \textbfit{motion primitive graph} $\mathcal{G}$ is a directed graph whose nodes are fixed, periodic or transient motion primitives and edges are Class 1 or Class 2 transitions. The motion primitive graph consists of $N$ motion primitive nodes $\{\motionprimitive_1,\ldots,\motionprimitive_N\}$ with transitions $\transitions_{ij}$ potentially connecting node $\motionprimitive_i$ to node $\motionprimitive_j$ for ${i,j \in \{1, \ldots, N\}}$.
\end{definition}




\begin{figure}
     \begin{subfigure}[b]{0.5\textwidth}
      \centering
      \begin{tikzpicture}
      \node[fixed node] (Stand) at (0,0) {\small Stand};
      \node[periodic node] (Walk) at (4,0) {\small Walk};
      \end{tikzpicture}
      \caption{Original incomplete graph, no transition}
     \end{subfigure} \\[6pt]
     \begin{subfigure}[b]{0.5\textwidth}
      \centering
      \begin{tikzpicture}
      \node[fixed node] (Stand) at (0,0) {\small Stand};
      \node[periodic node] (Step) at (2,0) {\small Step};
      \node[periodic node] (Walk) at (4,0) {\small Walk};
      \draw[class1] (Stand) -- node[] {} (Step);
      \draw[class2] (Step) -- node[] {} (Walk);
      \end{tikzpicture}
      \caption{Periodic Step-in-Place intermediate primitive}
     \end{subfigure} \\[6pt]
     \begin{subfigure}[b]{0.5\textwidth}
      \centering
      \begin{tikzpicture}
      \node[fixed node] (Stand) at (0,0) {\small Stand};
      \node[transient node, scale=0.75] (Accel) at (2,0) {\small Accelerate};
      \node[periodic node] (Walk) at (4,0) {\small Walk};
      \draw[class1] (Stand) -- node[] {} (Accel);
      \draw[class2] (Accel) -- node[] {} (Walk);
      \end{tikzpicture}
      \caption{Transient Acceleration intermediate primitive}
     \end{subfigure} \\[6pt]
     \begin{subfigure}[b]{0.5\textwidth}
      \centering
      \begin{tikzpicture}
      \node[fixed node] (Stand) at (0,0) {\small Stand};
      \node[periodic node,align=center] (Walk) at (4,0) {\footnotesize Safe \\ \footnotesize Walk};
      \draw[class1] (Stand) -- node[] {} (Walk);
      \end{tikzpicture}
      \caption{Modification of existing Walk primitive}
     \end{subfigure}
   \caption{Examples for mending an incomplete graph between Stand and Walk motion primitives.
   }
   \label{stand-to-walk}
   \vspace{-5mm}
\end{figure}
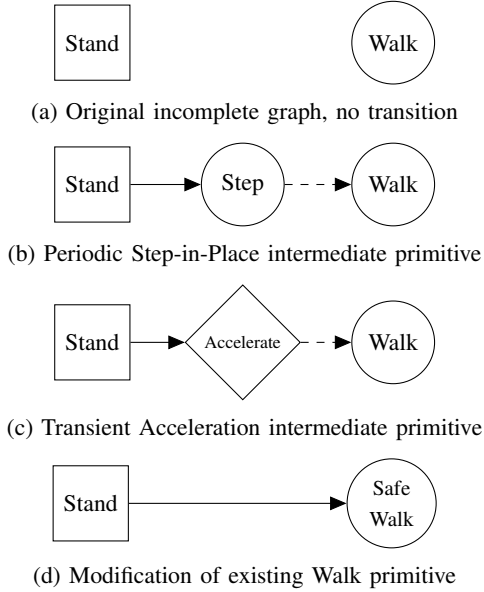

An example of a motion primitive graph can be seen in Figure \ref{quad_graph}.
In general, there are no guarantees that a graph from an arbitrary set of primitives is connected, although it is a necessity for each primitive to be reachable.
While ensuring a collection of motion primitives forms a connected graph is outside the scope of this paper, we posit this can be done by the modification of existing nodes or introduction of intermediate nodes.
This is illustrated in Figure \ref{stand-to-walk} with an example using Stand and Walk primitives for a legged robotic system.
Initially, we have no transition between the Stand and Walk primitives, perhaps due to the violation of a no-slip safety constraint.
Examples to make this graph connected include: the introduction of a periodic Step-in-Place primitive, the introduction of a transient Accelerate primitive, and modifying the Walk primitive to respect the no-slip safety constraint.
While this is a significant topic in its own right, we will not investigate this further in this paper and we make the assumption the given nodes form a connected graph.


\section{Constructing the Motion Primitive Graph}
\label{sec:graph}

We now turn to the major contribution of this paper:
verifying the existence of safe transitions between motion primitives. For the subsequent
analysis, consider prior primitive $A$ and posterior primitive $B$. 


\subsection{Transitions via Safety Oracle}
\label{subsec:safety_oracle}
Clearly, the goal is to determine if a
transition exists from
$A$ to $B$ based on the condition ${x_A^*(t_A) \in \mathcal{S}_B(t_B)}$ in Definition~\ref{def:transition}.
The setpoint $x_A^*(t)$ is known from the construction of motion primitive $A$, and if we
have a formulation for the safe RoA $\mathcal{S}_B(t)$, the desired result follows.
Note that the safe set $\mathcal{C}_B(t)$ is designer-prescribed and known, and the remaining component of
$\mathcal{S}_B(t)$ is the region of attraction. 
Because determining the RoA explicitly is difficult, if not impossible, for general systems and primitives we require an alternative method to determine if a transition
exists.
We make two observations: 
\begin{enumerate}
\item We do not necessarily need an explicit formulation for $\mathcal{S}_B(t)$ to determine if $x_A^*(t_A) \in \mathcal{S}_B(t_B)$.
\item There is at least one point we know is in $\mathcal{S}_B(t)$: $x_B^*(t)$. 
\end{enumerate}
From observation 2, we hypothesize that there exists a neighborhood $\mathcal{E}_B(t)$ of
$x_B^*(t)$, called \textit{explicit safe region of attraction} (see Definition~\ref{def:primitive}), that is a subset of $\mathcal{S}_B(t)$ and is tractable to compute.
This can be constructed based on the safe set and a \textit{conservative} estimate of the RoA. For example, if the control law $k_B$ renders $x^*_B$ exponentially stable, the converse Lyapunov theorem can be used to estimate the RoA via a norm bound.
The existence of the explicit safe RoA $\mathcal{E}_B(t)$ combined with observation 1 leads to the idea of the \textit{safety oracle}.
\begin{definition}{}
For motion primitive $\motionprimitive$,
a \textbfit{safety oracle} with horizon ${T>0}$ is a map $\mathcal{O}: \mathcal{X} \times \mathbb{R} \rightarrow \{0,1\}$
defined by:
\begin{align*}
\mathcal{O}(x_0,t_0) = 
\begin{cases} 
      1 & \text{if } \phi_t(x_0) \in \mathcal{C}(t),  \quad \forall t \in [0,T], \\
      & \quad \text{and } \phi_T(x_0) \in \mathcal{E}(t_0+T), \\
      0 & \text{otherwise}.
\end{cases}
\end{align*}
\end{definition}


This definition reduces the verification of safe transitions to the evaluation of the safety oracle $\mathcal{O}$ as follows.
\begin{theorem}
\label{thm}
\textit{
If ${\exists t_B \in \mathbb{R}}$ such that ${\mathcal{O}_B(x_A^*(t_A),t_B) = 1}$, then there exists a transition $\transitions_{AB}$ starting at $t_A$ from motion primitive A to motion primitive B.
}
\end{theorem}
\proof
By definition, ${\mathcal{O}_B(x_A^*(t_A),t_B) = 1}$ implies
\begin{align}
\begin{split}
& \phi_t^B(x_A^*(t_A)) \in \mathcal{C}_B(t), \quad \forall t \in [0,T], \\
& \phi_T^B(x_A^*(t_A)) \in
\mathcal{E}_B(t_B+T) \subseteq
\mathcal{S}_B(t_B+T),
\end{split}
\label{eq:proof_step1}
\end{align}
where $\mathcal{E}_B$ is a subset of $\mathcal{S}_B$ by construction.
By the definition~(\ref{eq:safeRoA}) of $\mathcal{S}_B$,
the second line of~(\ref{eq:proof_step1})
implies
\begin{equation}
\phi_t^B(x_A^*(t_A)) \in \mathcal{C}_B(t), \quad \forall t \geq T
\label{eq:proof_step2}
\end{equation}
and
\begin{equation*}
\lim_{\theta \to \infty} \phi_{\theta}^B \big( \phi_T^B(x_A^*(t_A)) \big) - x_B^*(\theta + T + t_B) = 0.
\end{equation*}
Since ${\phi_{\theta}^B \big( \phi_T^B(x_A^*(t_A)) \big) = \phi_{\theta+T}^B(x_A^*(t_A))}$, a change of coordinates ${t=\theta+T}$ leads to form~(\ref{eq:RoA}) and gives ${x_A^*(t_A) \in \Omega_B(t_B)}$.
With the first line of~(\ref{eq:proof_step1}) and~(\ref{eq:proof_step2}), this leads to ${x_A^*(t_A) \in \mathcal{S}_B(t_B)}$, which proves the existence of transition $\mathcal{T}_{AB}$.
\hfill $\blacksquare$

\begin{figure}
   \centering
   \includegraphics[width=0.85\columnwidth]{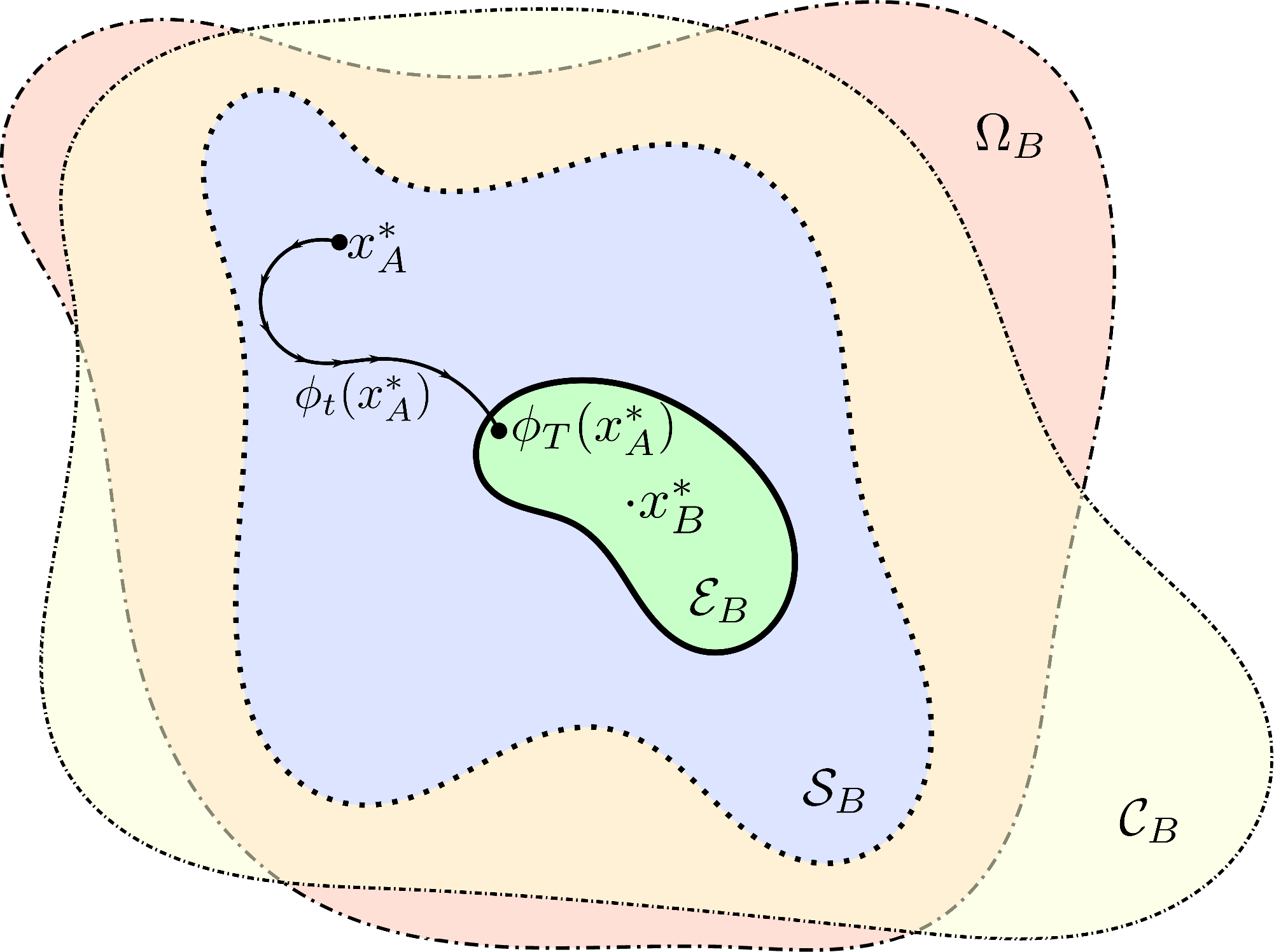}
   \caption{Illustration of the relationship of motion primitives $A$ and $B$ by the setpoints $x_A^*$ and $x_B^*$, flow $\phi_t(x_A^*)$, region of attraction $\Omega_B$, safe set $\mathcal{C}_B$, implicit safe RoA $\mathcal{S}_B$, and explicit safe RoA $\mathcal{E}_B$.}
   \label{blob-chart-explicit}
   \vspace{-4mm}
\end{figure}

Theorem~\ref{thm} provides a practical way of verifying the existence of transitions: one needs to evaluate ${\mathcal{O}_B(x_A^*(t_A),t_B)}$ for various $t_B$ values.
When evaluating $\mathcal{O}_B$, the flow $\phi_t(x_A^*(t_A))$ over ${t \in [0,T]}$ can be computed by numerical simulation of~(\ref{eq:CL_dynamics}) which can be terminated if the flow enters $\mathcal{E}_B$.
This is illustrated in Figure~\ref{blob-chart-explicit}. While the method is computationally tractable for finite $T$, it is conservative in the sense that Theorem~\ref{thm} involves a sufficient condition and ${\mathcal{O}_B(x_A^*(t_A),t_B)=0}$ does not necessarily rule out the existence of a transition.
However, this conservativeness becomes negligible if $T$ is large enough, since the flow from all points of $S_B$ eventually enter $\mathcal{E}_B$.

\subsection{Constructing the Graph}
Up until this point, we referred to motion primitives with distinct setpoints $x^*(t)$ and safe regions of attraction $\mathcal{S}(t)$.
In practice, these may depend on certain parameters; dynamic behaviors may take a range of continuous or discrete arguments, such as walking speed or body orientation
while standing in the case of legged robots.
To support this with our framework, we quantize the argument space of behaviors, and treat each quanta as a unique motion primitive when constructing the motion primitive graph.

For each ordered pair of primitive, $A$ to $B$, we can then apply the safety oracle with respect to a discretization of time for primitive $A$. The flow in the safety oracle is computed via numerical integration of the system under $k_B$ for a parameterized time $T$. If the safety oracle returns True for all times in the discretization, we add a Class 1 transition to the graph from $A$ to $B$. If it returns True for only some times, we add a Class 2 transition to the graph. If the safety oracle never returns True, then no transition is added. Details can be found in Algorithm~\ref{algo:motGraph}.
It is important to note that this process can be done offline---in fact, 
this procedure only needs to be repeated if the dynamic model changes, existing primitives are modified, or new primitives are added.
In the latter two cases, only a subset of motion primitive pairs needs to be rechecked for transitions.
This results in a tractable procedure for constructing the motion primitive
graph.

\begin{algorithm}[tb]
\caption{\motGraph}
\label{algo:motGraph}
\begin{small}
\begin{algorithmic}[1]
\Function{\motGraph ($\motionprimitive$,T)}{}
    \State $\transitions = \emptyset$\Comment{$\transitions$ List of edges for motion primitive graph $\mathcal{G}$}
  \For{$A, B\in \motionprimitive$}
        \State set $\transitions_{AB} = \emptyset$ \Comment{Assume no transition exists}
        \State set $SO{(\cdot,\cdot)}$ = False \Comment{Safety Oracle responses}
        \For{$(t_A,t_B) \in [t_{0A},t_{{\rm f}A}]\times [t_{0B},t_{{\rm f}B}]$}
          \State set $SO{(t_A,t_B)} = ${\textsc{SafetyOracle}$(f_{{\rm cl},B(t_B)},x^*_A,$\WRP $ \mathcal{E}_B(t_B), \mathcal{C}_B(t_B), T)$}
         \EndFor
        \If{ any($SO{(\cdot, \cdot)} =$ True)}
          \State set $\mathcal{T}_{AB} =$ Class 2
        \EndIf
        \If{$\forall\ t_A$ any($SO{(t_A,\cdot)}$ = True)}
          \State set $\mathcal{T}_{AB} =$ Class 1
        \EndIf
        \State \textsc{AddEdge}$(\mathcal{T}, \transitions_{AB})$
    \EndFor
    \State \Return $\mathcal{G} = (\mathcal{P},\transitions)$
\EndFunction

\Statex
\Function{SafetyOracle}{$f_{\rm cl},x^*$, $\mathcal{E}, \mathcal{C}, T$}
  \State $t = 0$
  \While{$t < T$}
    \State $\phi_t(x^*) =$ \textsc{Integrate}($f_{\rm cl}(x^*)$)
    \If{$\phi_t(x^*) \in \mathcal{E}$}
       \State \Return True
    \EndIf
    \If{$\phi_t(x^*) \notin \mathcal{C}$}
       \State \Return False
    \EndIf
  \EndWhile
  \State \Return False
\EndFunction
\end{algorithmic}
\end{small}
 \label{alg:buildmotionprimitivegraph}
 \end{algorithm}
\vspace{-15mm}
\section{Application to Quadrupeds}
\label{sec:quadruped}

The proposed concepts are applied to the Unitree A1 quadrupedal robot, shown in
Figure~\ref{quad_graph}, with a small set of motion primitives. 
In this setting, we define the local configuration space coordinates as $q\in \mathcal{Q}\subset \mathbb{R}^n$ with the full state space given by $x=(q,\dot q)\in \mathcal{X}= T\mathcal{Q} \subset \mathbb{R}^{2n}$; in the case of the Unitree A1, $n=18$. The control input is given by $u\in \mathcal{U} \subset \mathbb{R}^{m}$, with $m=12$ actuators for the A1. During walking, the no slip condition of the feet is enforced via \textit{holonomic constraints}, encoded by $c(q)\equiv 0$ for $c(q)\in \mathbb{R}^{n_c}$, where $n_c$ depends on the number of feet in contact with the ground. Differentiating $c(q)$ twice, D'Alembert's principle applied to the constrained Euler-Lagrange equations gives:
\begin{align*}
    D(q)\ddot{q} + H(q,\dot{q}) &= Bu + J(q)^\top \lambda, \\ 
    J(q)\ddot{q} + \dot{J}(q,\dot{q})\dot{q} &= 0, 
\end{align*}
where $D(q)\in\mathbb{R}^{n\times n}$ is the mass-inertia matrix, $H(q,\dot q)\in\mathbb{R}^n$ contains the Coriolis and gravity terms, $B\in\mathbb{R}^{n\times m}$ is the actuation matrix, $J(q) = \partial c(q)/\partial q \in \mathbb{R}^{n_c \times n}$ is the Jacobian of the holonomic constraints, and $\lambda\in \mathbb{R}^{n_c}$ is the constraint wrench. This can be converted to control affine form:
\begin{align*}
    \dot{x} = \underbrace{\begin{bmatrix}\dot{q}\\-D(q)^{-1}(H(q,\dot q)-J(q)^\top \lambda)\end{bmatrix}}_{f(x)} + \underbrace{\begin{bmatrix}0\\D(q)^{-1}B\end{bmatrix}}_{g(x)}u,
\end{align*}
where the mappings ${f:\mathcal{X}\to \mathbb{R}^n}$ and ${g:\mathcal{X}\to\mathbb{R}^{n\times m}}$ are assumed to be locally Lipschitz continuous. 
In the case of legged locomotion, the alternating sequences of continuous and discrete dynamics are captured within the \textit{hybrid dynamics} framework. For the sake of simplicity, and because only the continuous dynamics are considered in the controller design process, this development will be omitted; please refer to \cite{westervelt2018feedback} for more details. 
Finally, the outputs being driven to zero by the controllers in each respective motion primitive can be represented as:
\begin{align*}
    y(x,t) = y^{\rm a}(x) - y^{\rm d}(t),
\end{align*}
where the actual measured state $y^{\rm a}:\mathcal{X}\to \mathbb{R}^o$ and the desired state $y^{\rm d}:\mathbb{R}\to\mathbb{R}^o$ are smooth functions.

\subsection{Quadruped Motion Primitives Preliminaries}
Recall that our framework is agnostic to the implementation details of the primitives
themselves, granted the assumptions in Section \ref{sec:preliminaries} are met. Nevertheless, the implementation details, along with the computable components of the motion primitive tuple, $(x^*, k, \mathcal{C}, \mathcal{E})\subset \motionprimitive$, are elucidated for each primitive in our application. We begin by defining some common attributes.

\begin{primitive}{Common Safe Sets}
For all primitives, the safety functions employed on hardware include joint position and velocity limits,
whereby a combined safe set can be constructed as:
\begin{align*}
\mathcal{C}_{q,\dot{q}} = \{x \in \mathcal{X} : \;&  h_{q_{\rm min}}(x) = q - q_{\rm min} \geq 0,\\
            & h_{q_{\rm max}}(x)= q_{\rm max} - q \geq 0,\\
            & h_{\dot{q}_{\rm min}}(x) = \dot{q} - \dot{q}_{\rm min} \geq 0,\\
            & h_{\dot{q}_{\rm max}}(x) = \dot{q}_{\rm max} - \dot{q} \geq 0
            \}.
\end{align*}
Additionally, many primitives require immobile ground contact with all feet for safety. Assuming a flat ground at $z=0$, this can be captured by safety functions:
\begin{align*}
\mathcal{C}_{\rm ground} = \{x \in \mathcal{C}_{q,\dot{q}} : \;&  h_{{\rm g}1}(x) = -fk_z(q) \geq 0, \\
&  h_{{\rm g}2}(x) = -J(q)\dot{q} = 0
            \},
\end{align*}
where $fk_z:\mathcal{Q}\to\mathbb{R}^{4}$ is the $z$-component of the forward kinematics of the feet of the quadruped.
\end{primitive}
\begin{primitive}{Common Explicit Safe Regions of Attraction}
As discussed in Section~\ref{subsec:safety_oracle}, we can choose a conservative safe RoA estimate $\mathcal{E}$ and compensate with an increased integration horizon T. For all the primitives in the experiment, we take a conservative norm bound about $x^*(t)$:
%
\begin{align*}
\mathcal{E}(t) = \{x\in \mathcal{X} : ||x-x^*(t)|| < r\},
\end{align*}
where the value of $r$ is determined empirically for each primitive.
\end{primitive}
\begin{primitive}{Motion Profiling}
Where indicated, motion primitives utilize cubic polynomial motion profiling over a fixed duration ${\Theta>0}$. This is computed via linear matrix equation:
\begin{align*}
    \begin{bmatrix} c_0 \\ c_1 \\ c_2 \\ c_3 \end{bmatrix} =
    \begin{bmatrix}
       1 & 0 & 0 & 0 \\ 
       1 & \Theta & \Theta^2 & \Theta^3 \\ 
       0 & 1 & 0 & 0 \\ 
       0 & 1 & 2\Theta^2 & 3\Theta^2 \\ 
    \end{bmatrix}^{-1}
    \begin{bmatrix} q_0 \\ q_{\rm f} \\ \dot{q}_0 \\ \dot{q}_{\rm f} \end{bmatrix},
\end{align*}
where ${x_A^* = (q_0,\dot q_0)}$ and ${x^*_B = (q_{\rm f},\dot q_{\rm f})}$. Taking the desired state as ${y^{\rm d}(t) = (q^{\rm d}(t),\dot{q}^{\rm d}(t))}$, we have the polynomial:
\begin{align}
y^{\rm d}(t) &= \label{eq:mp}
\begin{cases} 
      (q_0, \dot q_0) & \text{if } t < 0, \\
      \begin{matrix}(c_0 +c_1t+c_2t^2+c_3t^3, \\ c_1+2c_2t+3c_3t^)\end{matrix}  &\text{if } t\in[0,\Theta], \\
      (q_{\rm f},\dot{q}_{\rm f}) & \text{if } t > \Theta. \\
\end{cases} 
\end{align}
\end{primitive}

\subsection{Quadruped Motion Primitives}
We considered the motion primitives listed below:

\begin{primitive}{Lie}
The \textit{Lie} is a fixed motion primitive that rests the quadruped on the ground with the
legs in a prescribed position. The feedback controller is a joint-space PD controller:
\begin{align}
    k(x,t) = - K_P y(x,t) - K_D\dot y(x,t), \label{eq:PD}
\end{align}
where $y^{\rm d}(x,t)$ is the previously explained cubic spline in \eqref{eq:mp} from the pose when the primitive is first applied to the Lie goal pose, $x^*_{\rm Lie}$. The safe set for Lie requires ground contact from all four feet:
%
$\mathcal{C}_{\rm Lie} = \mathcal{C}_{\rm ground}$.
\end{primitive}

\begin{primitive}{Stand}
For the \textit{Stand} fixed motion primitive, $x^*_{\rm Stand}$ prescribes the body to be at a specified height above the ground and the center of mass to lie above the centroid of the support polygon. Taking $\mathcal{X}_{\rm ext} = [\ddot q^\top, u^\top, \lambda^\top]^\top \in \mathbb{X}_{\rm ext}= \mathbb{R}^{18}\times \mathbb{R}^{12} \times \mathbb{R}^{n_c}$, this is done via an
Inverse-Dynamics Quadratic Program based controller (ID-QP):
\begin{ruledtable}
\vspace{-2mm}
{\textbf{\normalsize ID-QP:}}
\par\vspace{-5mm}{\small 
\begin{align}
        \label{eq:CLF-QP-d}
	{\mathcal{X}_{\rm ext}}^* = \underset{\mathcal{X}_{\rm ext}\in\mathbb{X}_{\rm ext}}{\mathrm{argmin}} &\hspace{3mm} \|J_y(q,\dot q)\ddot{q} + \dot J_y(q,\dot{q})\dot q - \tau(x,t) \|^2 + \sigma W(\mathcal{X}) \notag \\
	\mathrm{s.t.} 		&\hspace{3mm} D(q)\ddot q + H(q,\dot q) = Bu + J(q)^\top \lambda \notag \\
                		&\hspace{3mm} J(q)\ddot{q} + \dot{J}(q,\dot{q})\dot{q} = 0 \notag \\
                		&\hspace{3mm} u\in \mathcal{U}\notag  \\
                		&\hspace{3mm} \lambda\in\mathcal{FC}(x)\notag  
\end{align}}\vspace{-7mm}\par
\end{ruledtable}
\noindent where $\mathcal{U}=[-33.5, 33.5]^{12}$ is limited by the available torque at each motor, $\sigma$ and $\mathcal{W}$ are regularization terms for numerical stability of the QP, and $\mathcal{FC}(x)\in\mathbb{R}^{n_c}$ is a friction cone condition to enforce no slipping of the feet. 
The motion profile \eqref{eq:mp} in center of mass task-space is used with a PD control law to produce $\tau(x,t)$ in the objective function:
\begin{align*}
    \tau(x,t) = K_P y(x,t) + K_D\dot y(x,t).
\end{align*}
Finally, $k(x,t) = u^*$. 
We have the safe set: $\mathcal{C}_{\rm Stand} = \mathcal{C}_{\rm ground}.$
\end{primitive}

\begin{primitive}{Walk}
The \textit{Walk} periodic motion primitive locomotes the quadruped forward by tracking a stable walking
gait.  Stable walking gaits were generated using a biped-decomposition technique as described in
\cite{ma2021coupled} and combined in a gait library, \cite{reher_traj_lib2}. Specifically, the desired output is written as:
\begin{align*}
    x^{*}(t) =(x^{*}_B(t), \mathcal{R}x^{*}_B(t)),
\end{align*}
where $x^{*}_B(t)$ is the desired behavior generated by the NLP toolbox FROST \cite{hereid2017frost} for one biped, and ${\mathcal{R}\in \mathbb{R}^{o\times o}}$ is a mirroring matrix relating the behavior of the two subsystems.
The gaits are tracked using the joint-space PD
control law in \eqref{eq:PD}. This primitive can be called with an argument to control forward speed, specifically: Walk(in~place), Walk(slow), Walk(medium) and Walk(fast).
Any slipping of stance feet is deemed unsafe, which is described by the safety function:
\begin{align*}
h_i(x) = \begin{cases}
          -J^i_{xy}(q)\dot{q} &\text{if } fk^i_z(q) \leq 0, \\
          0 &\text{otherwise},
       \end{cases} \\
\mathcal{C}_{\rm Walk} = \{x \in \mathcal{C}_{q,\dot{q}} : \;h_i(x) = 0, i \in \mathcal{F}&
            \} ,
\end{align*}
where $fk^i_z(q):\mathcal{Q}\to\mathbb{R}$ is the forward kinematics $z$-component, $J^i_{xy}(q):\mathcal{Q}\to\mathbb{R}^{2\times n}$ is the $x,y$-components of the Jacobian for each foot, and $\mathcal{F}\subset\{1,...,4\}$ is an index set of feet in ground contact.
\end{primitive}

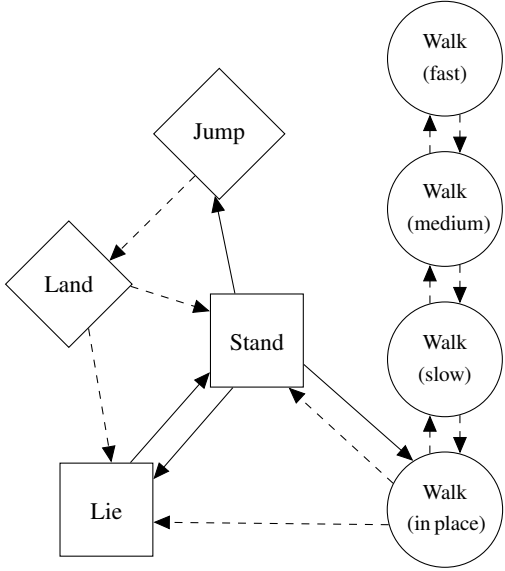
\begin{figure}
   \centering
    \begin{tikzpicture}
      \node[fixed node,text height=0.63cm,text width=1cm, align=center] (Stand) at (0,-0.75) {\small Stand\vspace{0.37cm}};

      \node[fixed node,text height=0.63cm,text width=1cm, align=center] (Lie) at (-2,-3) {\small Lie\vspace{0.37cm}};

      \node[periodic node,text width=1cm, align=center] (Walk) at (2.5,-3) {\footnotesize Walk \\
\footnotesize(in\,place)};
      \node[periodic node,text width=1cm, align=center] (Walk1) at (2.5,-1) {\footnotesize Walk \\ \footnotesize (slow)};
      \node[periodic node,text width=1cm, align=center] (Walk2) at (2.5,1) {\footnotesize Walk \\ \footnotesize (medium)};
      \node[periodic node,text width=1cm, align=center] (Walk3) at (2.5,3) {\footnotesize Walk \\ \footnotesize (fast)};

      \transition[class1](Stand,Lie)();
      \transition[class1](Lie,Stand)();
      \transition[class1](Stand,Walk)();
      \transition[class2](Walk,Stand)();

      \transition[class2](Walk,Walk1)();
      \transition[class2](Walk1,Walk)();
      \transition[class2](Walk1,Walk2)();
      \transition[class2](Walk2,Walk1)();
      \transition[class2](Walk2,Walk3)();
      \transition[class2](Walk3,Walk2)();

      \node[transient node,text width=1cm, align=center] (Land) at (-2.5,0) {\small Land};
      \node[transient node,text width=1cm, align=center] (Jump) at (-0.5,2) {\small Jump};
      \transition[class1](Stand,Jump)();
      \transition[class2](Jump,Land)();
      \transition[class2](Land,Stand)();
      \transition[class2](Land,Lie)();
      \transition[class2](Walk,Lie)();
    \end{tikzpicture}
   \caption{Computed motion primitive graph for the experimental set of quadruped motion primitives.
   }
   \label{real_quad_graph}
   \vspace{-3.5mm}
\end{figure}

\vspace{-3mm}

\begin{primitive}{Jump}
The transient \textit{Jump} primitive is a four-legged vertical jump, performed by task-space tracking of $x^*_{\rm Jump}$, a
prescribed center of mass trajectory to reach a specified takeoff velocity. The trajectory is generated via \eqref{eq:mp} and does not take constraints (input or otherwise) into account.
The trajectory is tracked via a task-space PD control law:
\begin{align*}
    k(x,t) = J_y(x)^\top\big(- K_P y(x,t) - K_D\dot y(x,t)\big),
\end{align*}
where $J_y:\mathcal{X}\to \mathbb{R}^{o\times m}$ is the Jacobian of the outputs. Like other transient
primitives, this requires a ``next primitive'' argument. For the purpose of this experiment, the
next primitive will always be Land.
To be in the safe set, the Jump primitive requires all feet to be in contact at ${t=0}$, and no feet to be in contact after prescribed takeoff time.
For the aerial phase, we define:
\begin{align*}
\mathcal{C}_{\rm aerial} = \{x \in \mathcal{C}_{q,\dot{q}} : \;&  h(x) = fk_z(q) \geq 0 \}.
\end{align*}
The composite safe set is then:
\begin{align*}
\mathcal{C}_{\rm Jump}(t) &= 
\begin{cases} 
      \mathcal{C}_{\rm ground} & \text{if } t < T_{\rm t}, \\
      \mathcal{C}_{q,\dot{q}} & \text{if }  T_{\rm t} \leq t < T_{\rm t} + \epsilon , \\
      \mathcal{C}_{\rm aerial} & \text{if }  T_{\rm t} + \epsilon \leq t < T_{\rm f}, \\
\end{cases} 
\end{align*}
where $T_{\rm t}$ and $T_{\rm f}$ are the takeoff time and final time prescribed by the trajectory, and ${\epsilon>0}$ allows for non-simultaneous contact in the safe set.
\end{primitive}
 
\begin{primitive}{Land}
The \textit{Land} primitive is a high-damping joint-space PD control law, as in \eqref{eq:PD}, about a crouched pose, $x^*_{\rm Land}$, to cushion the quad as the
legs make contact with the ground from an airborne state. In this experiment, the next primitive
will always be Stand. The safe set for Land requires all feet to not be in contact at $t=0$ 
and all feet to reach contact within a small time window after initial ground contact:
\begin{align*}
\mathcal{C}_{\rm Land}(t) &= 
\begin{cases} 
      \mathcal{C}_{\rm aerial} & \text{if } t < T_{\rm c}, \\
      \mathcal{C}_{q,\dot{q}} & \text{if }  T_{\rm c} \leq t < T_{\rm c} + \epsilon,  \\
      \mathcal{C}_{\rm ground} & \text{if }  T_{\rm c} + \epsilon \leq t, \\
\end{cases} 
\end{align*}
where $T_{\rm c}$ is the time of first ground contact.
\end{primitive}


\begin{figure}
    \centering
    \includegraphics[width=\linewidth]{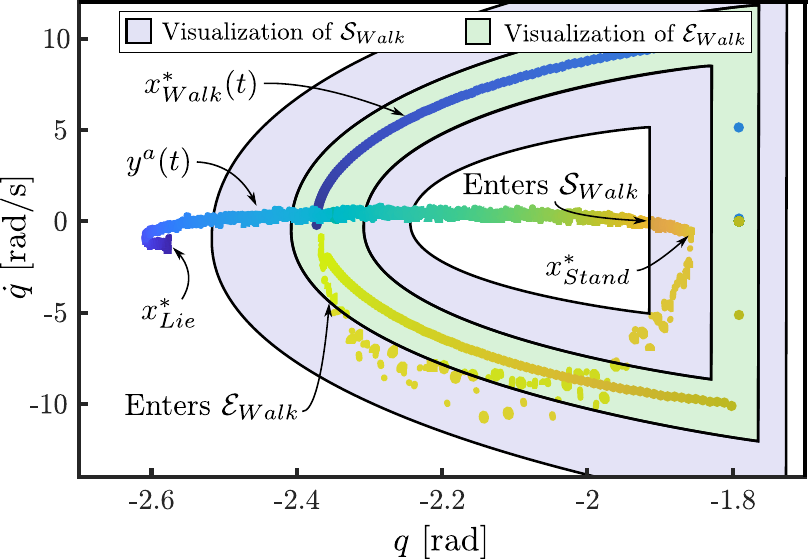}
    \caption{Depiction of the transition from Lie to Stand to Walk(in place) during the hardware experiments. The inclusion of $x^*_{\rm Stand}$ in the approximate $\mathcal{S}_{\rm Walk}$ can be seen, and upon switching to the controller for the walk primitive, the flow of the system eventually reaches $\mathcal{E}_{\rm Walk}$.}
    \label{fig:hardware_transition}
    \vspace{-5mm}
\end{figure}

\begin{figure*}
  \includegraphics[width=\textwidth]{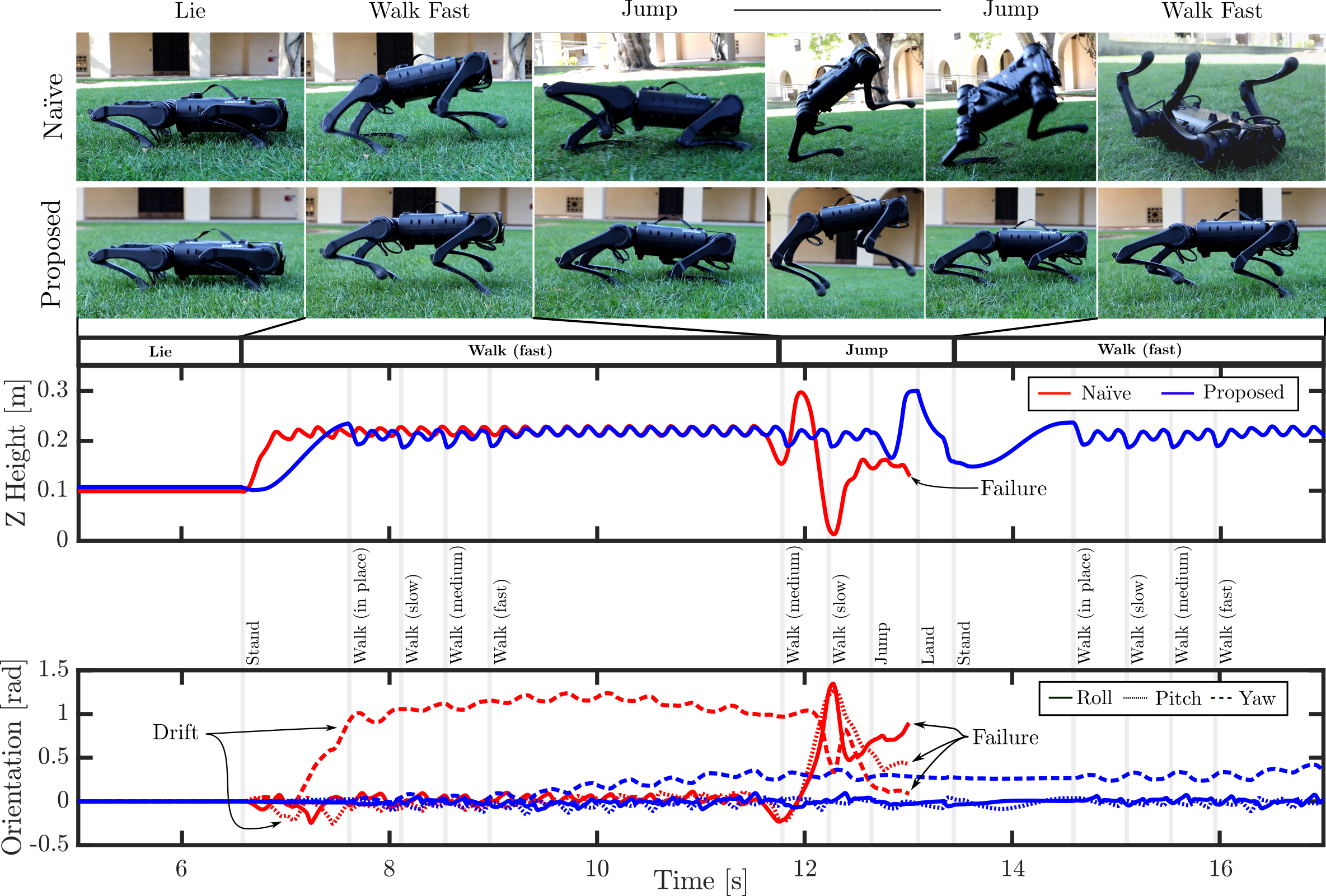}
  \caption{Motion primitive graph traversal for a sample sequence of desired primitives on quadrupedal system. \textit{Top:} Gait tiles comparing the proposed method to a na\"ive method, which results in a violation of safety. \textit{Bottom:} center of mass height and orientation comparison between na\"ive transitions and the
proposed graph-traversal transitions. Blocks under the upper plot show desired behavior and vertical lines with corresponding text indicate the mode used by the proposed method. Note the substantial drift and ultimate failure when using the na\"ive transitions.}
  \label{comparison_snapshot}
  \vspace{-3mm}
\end{figure*}

\vspace{-2mm}
\subsection{Implementation and Experimental Results}
Each individual primitive was implemented as a C++ module with function calls for the 
computable portions of the motion primitive tuple, $(x^*, k, \mathcal{C}, \mathcal{E})\subset \motionprimitive$. A C++ implementation of Algorithm~\ref{algo:motGraph} was applied, utilizing the
Pinocchio Rigid Body dynamics library \cite{pinocchioweb} and Boost's odeint with a
\texttt{runge\_kutta\_cash\_karp54} integration scheme \cite{BoostLibrary}. Note that in the construction of the motion primitive graph,
the hybrid nature of the system was taken into account.

The resulting motion primitive graph is visualized in Figure \ref{real_quad_graph}. The graph
shows that we can freely transition from Lie to Stand, but must go through Walk(in~place)
before reaching other walk speeds. Additionally, the Class 2 transitions from the
Walk states indicates the time-dependent nature of the transition: we can only leave the Walk
primitive at certain times in the orbit.
\begin{table}[b]
\vspace{-5mm}
\caption{Test Sequence of Goal Motion Primitives}
\label{sequence}
\begin{center}
\resizebox{\linewidth}{!}{
\begin{tabular}{|l|l|l|l|l|l|}
\hline
\textbf{Time} & 0 s & 3 s & 8 s & 11 s & 16 s \\
\hline
\shortstack[l]{\textbf{Goal} \\ \textbf{Primitive}} & Lie & Walk(fast) & Jump & Walk(fast) & Lie\\
\hline
\shortstack[l]{\textbf{Inter.} \\ \textbf{Primitives}} & --- & \shortstack[l]{\textbf{Lie} \\ Stand \\ Walk(in\,place) \\ Walk(slow) \\ Walk(medium) \\ \textbf{Walk(fast)}} & \shortstack[l]{\textbf{Walk(fast)} \\ Walk(medium) \\ Walk(slow) \\ Walk(in\,place) \\ Stand\\ \textbf{Jump}} & \shortstack[l]{\textbf{Jump} \\ Land \\ Stand \\ Walk(in\,place) \\ Walk(slow) \\ Walk(medium) \\ \textbf{Walk(fast)}} & \shortstack[l]{\textbf{Walk(fast)} \\ Walk(medium) \\ Walk(slow) \\ Walk(in\,place) \\ \textbf{Lie}} \\
\hline
\end{tabular}}
\end{center}
\vspace{-2mm}
\end{table}
Using the motion primitive graph, offline depth-first search was used to compute the path
from each known current primitive to a desired motion primitive. The solutions were implemented as
an online lookup table.




To validate the method experimentally, a sequence of goal primitives was built to
exercise the different transition types in the quadruped example. The sequence is shown in
Table \ref{sequence}. Each desired primitive in the sequence was commanded at the listed time.
Two cases were run with the sequence: the first with na\"ive, immediate application of the goal
motion primitive, and the second with our proposed traversal algorithm executing intermediate
motion primitives based on the motion primitive graph.



With the proposed traversal algorithm executing intermediate primitives, we see the transition
from Lie to Walk(fast) first traverses through Stand, then Walk(in~place) and other walk speeds
until finally reaching Walk(fast) without violating the no-slip and other safety constraints.
The evolution of the hardware system with respect to the desired behavior as well as a representation of the 
implicit and explicit safe sets can be seen in Figure \ref{fig:hardware_transition}.
Likewise, when Jump is commanded, the current motion primitive
traverses downward through the walk speeds and then to Stand before executing the vertical
jump. As Jump is a transient primitive, this is followed by the Land and Stand primitives before
traversing back to Walk(fast). The sequence ends successfully by cycling back down through the
walk speeds to Walk(in~place) before finishing at Lie. Unlike the na\"ive transitions, the
graph traversal successfully and safely completes the sequence of desired motion primitives.

In the na\"ive case, the no-slip safety constraint is violated in the Lie to Walk(fast)
transition.  Furthermore, the Walk(fast) to Jump transition results in the failure of the Jump
primitive to track the desired behavior, as the Walk(fast) dynamic state at the transition time
is not in the region of attraction of the Jump primitive.  These results can be inferred from
the motion primitive graph, as there is no verification of a direct transition between Lie and
Walk(fast) and Walk(fast) and Jump. Ultimately, the direct transitions fail to maintain safety
and cannot perform the desired sequence of behaviors.

A video comparison between the na\"ive transitions and the graph-traversal transitions is
provided as an attachment (see \cite{Video}) and a snapshot comparison of gait tiles and collected data can be seen in Figure
\ref{comparison_snapshot}.

\section{CONCLUSIONS}
\label{sec:concl}

Motivated by the complex autonomy realized in the quasi-static robotics realm and heuristically on dynamic legged robots, this paper has 
leveraged ideas from dynamic systems to make the first steps toward formulating a theory of transition classes between
``motion primitives''. With these fundamentals, we proposed a tractable procedure to verify the existence of transitions between motion primitives through the use of a safety oracle and subsequently presented an algorithm for the construction of a ``motion primitive graph.'' This graph is used to determine transition paths via standard graph search algorithms.
To illustrate the viability of the method, it was applied to a quadrupedal robot and a set of quadrupedal motion primitives. This culminated in a demonstration of the capability of our method via experiments on hardware while the robot performed dynamic behaviors.  

There are a number of extensions that we intend to investigate in future. We
would like to expand the theory described in Section \ref{sec:graph} to address
real-world uncertainty and disturbances, and explore necessary conditions for
$\phi_T(x_A^*(t))$ to enter $\mathcal{E}_B(t)$ in this context. In systems with a large number of motion
primitives, the final graph may offer many paths from one primitive to another. We would like
to consider a ``weighted'' motion primitive graph, and understand what metrics are useful 
as weights for transitions or primitive executions (e.g. transition duration, peak torque,
etc.).
Additionally, we recognize that this method only addresses the nominal use case of
autonomy in dynamic systems, and does not offer a solution to arbitrary initial dynamic states
or disturbances that throw the state outside the safe region of attraction of the currently
active primitive. Future work intends to explore these open questions in the context of
primitive behaviors and the motion primitive graph.

\bibliographystyle{IEEEtran}
\bibliography{reference}	

\end{document}